\documentclass[conference]{IEEEtran}

\IEEEoverridecommandlockouts
\usepackage{cite}
\usepackage{amsmath,amssymb,amsfonts}
\usepackage{algorithmic}
\usepackage{algorithm}
\usepackage{graphicx}
\usepackage{textcomp}
\usepackage{xcolor}
\usepackage{booktabs}
\usepackage{academicons}
\usepackage{url}
\definecolor{orcidlogocol}{HTML}{A6CE39}
\usepackage{multirow}
\usepackage{tabularx}
\usepackage{makecell}
\usepackage{tabularx}
\usepackage{subcaption}
\usepackage{comment}
\usepackage{array}
\usepackage{threeparttable}

\usepackage{colortbl}
\usepackage{xcolor}
\usepackage{booktabs}
\usepackage{siunitx}
\definecolor{rowgreen}{RGB}{230,255,230} 

\begin{document}

\title{MedCoT-RAG: Causal Chain-of-Thought RAG for Medical Question Answering}

\author{
    Ziyu Wang\textsuperscript{*}, Elahe Khatibi\textsuperscript{*}, and Amir M. Rahmani \\
    University of California, Irvine, USA \\
    \{ziyuw31, ekhatibi, amirr1\}@uci.edu
    \thanks{\textsuperscript{*}Ziyu Wang and Elahe Khatibi contributed equally to this work.}
}

\maketitle

\begin{abstract}
Large language models (LLMs) have shown promise in medical question answering but often struggle with hallucinations and shallow reasoning, particularly in tasks requiring nuanced clinical understanding. Retrieval-augmented generation (RAG) offers a practical and privacy-preserving way to enhance LLMs with external medical knowledge. However, most existing approaches rely on surface-level semantic retrieval and lack the structured reasoning needed for clinical decision support. We introduce \textbf{MedCoT-RAG}, a domain-specific framework that combines causal-aware document retrieval with structured chain-of-thought prompting tailored to medical workflows. This design enables models to retrieve evidence aligned with diagnostic logic and generate step-by-step causal reasoning reflective of real-world clinical practice. Experiments on three diverse medical QA benchmarks show that MedCoT-RAG outperforms strong baselines by up to 10.3\% over vanilla RAG and 6.4\% over advanced domain-adapted methods, improving accuracy, interpretability, and consistency in complex medical tasks.
\end{abstract}


\section{Introduction}

Large language models (LLMs) have demonstrated strong performance across a wide range of medical NLP tasks, including question answering, clinical summarization, and diagnostic reasoning~\cite{xiong2024benchmarking, sohn2024rationale, wang2025healthq}. Their ability to synthesize biomedical knowledge from large-scale corpora has made them appealing as decision-support tools in healthcare. However, despite this progress, LLMs remain prone to hallucinations, inconsistencies, and reasoning failures—particularly when applied to complex clinical tasks that require causal understanding~\cite{pal2023med, khatibi2025cdf}. These challenges are exacerbated in safety-critical domains like medicine, where outputs must be not only accurate but also clinically plausible and interpretable.

Retrieval-augmented generation (RAG) has emerged as a promising approach to reduce hallucinations by grounding LLM outputs in relevant external knowledge~\cite{xiong2025rag, wu2024medical}. In medical settings, RAG provides a flexible, privacy-preserving way to enhance LLMs with domain-specific knowledge from biomedical literature, clinical guidelines, or structured EHRs—without the need for costly and often ineffective fine-tuning. However, most medical RAG systems face two core limitations. First, they rely on semantic similarity for retrieval, which often surfaces contextually similar but clinically irrelevant information~\cite{xiong2024benchmarking, cho2025k}. Second, they lack structured reasoning, producing unstructured answers that fail to reflect how clinicians generate differential diagnoses, assess treatments, or integrate evidence~\cite{wang2025healthq, wornow2024context}.

Clinical reasoning in real-world practice is structured and causal: physicians link symptoms to pathophysiology, evaluate differential diagnoses based on key features, and integrate evidence with patient-specific factors. These causal chains underpin real-world diagnostic reasoning but are largely missing from existing RAG pipelines, which focus on shallow context retrieval and unstructured generation. Even advanced medical RAG systems like MedRAG~\cite{zhao2025medrag} and those leveraging strong biomedical embeddings such as MedCPT~\cite{jin2023medcpt} or LinkBERT~\cite{yasunaga2022linkbert} retrieve context without explicitly modeling causal relationships. While chain-of-thought (CoT) prompting has improved reasoning in general LLMs~\cite{wei2022chain}, its use in medical settings often depends on generic templates that fail to capture the structured, diagnostic reasoning employed by clinicians.

\begin{figure*}[t]
    \centering
    \includegraphics[width=0.65\textwidth]{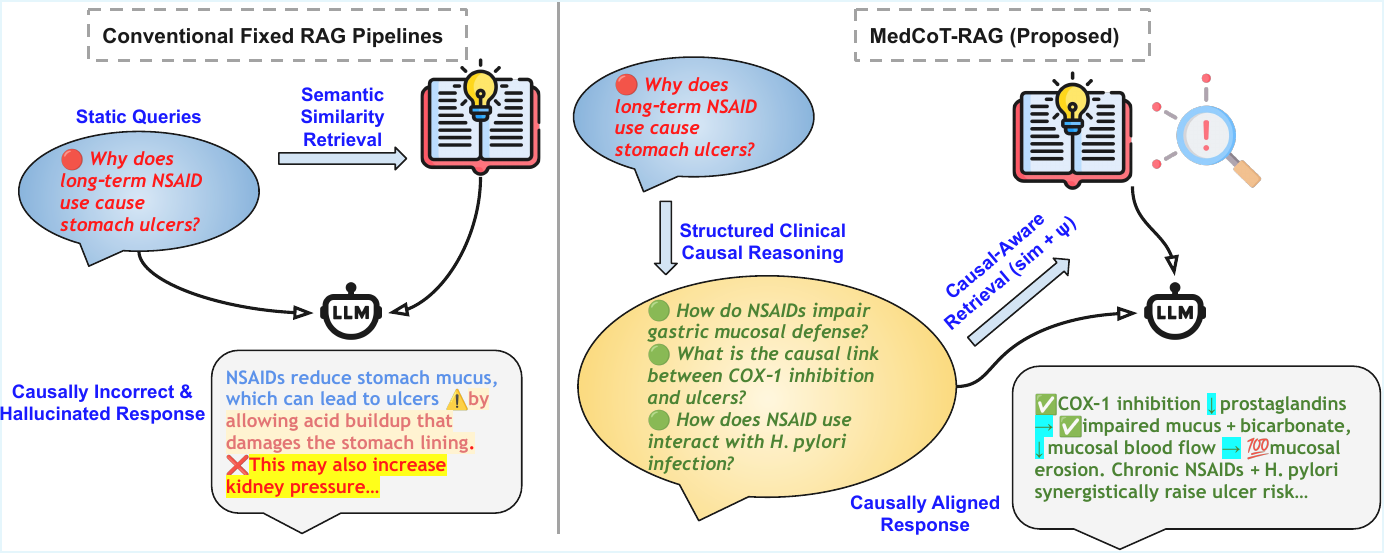}
    \caption{Comparison of conventional RAG and MedCoT-RAG. Conventional RAG uses static queries and semantic retrieval, often yielding hallucinated, causally incorrect responses. MedCoT-RAG combines causal-aware retrieval and structured reasoning via CoT splitting to produce causally consistent, clinically coherent answers.}
\label{fig:medcot-motivation}
\end{figure*}

In this work, we present \textbf{MedCoT-RAG}, a domain-specific RAG framework designed to improve medical question answering through structured causal reasoning and clinically grounded retrieval. MedCoT-RAG introduces a \textit{causal CoT prompting} strategy that guides LLMs through four stages aligned with clinical workflows: symptom analysis, causal pathophysiology, differential diagnosis, and evidence synthesis. Simultaneously, it performs \textit{causal-aware retrieval}, prioritizing documents that capture diagnostic reasoning, treatment mechanisms, and clinically relevant cause-effect relationships—beyond surface-level semantic similarity. This dual design enables MedCoT-RAG to generate medically coherent, interpretable, and reliable outputs, addressing key limitations of current biomedical RAG systems.

We evaluate \textbf{MedCoT-RAG} on three diverse and challenging medical QA benchmarks spanning various biomedical domains and question formats. Our evaluation includes strong baselines: a plain LLM without retrieval, vanilla RAG, RAG with MedCPT embeddings, and RAG with MedCPT combined with basic CoT prompting. We also compare against recent domain-specific retrieval-augmented methods, including RGAR~\cite{liang2025rgar} and RAG$^2$~\cite{sohn2024rationale}. Across all benchmarks, MedCoT-RAG consistently improves accuracy, reasoning quality, and clinical alignment. These results demonstrate the effectiveness of our structured causal reasoning framework and reinforce the importance of accuracy-based evaluation for complex medical question answering tasks.

\begin{itemize}
\item We propose MedCoT-RAG, a domain-specific RAG framework that combines causal-aware retrieval and structured CoT prompting to better capture clinical reasoning.
\item We design a novel retrieval scoring function that prioritizes documents based on causal medical relevance, including pathophysiological links, treatment rationales, and diagnostic mechanisms.
\item We demonstrate the effectiveness of MedCoT-RAG on three challenging medical QA benchmarks, showing consistent improvements in accuracy, reasoning depth, and clinical alignment over strong baselines.
\end{itemize}

\section{Method}

MedCoT-RAG is a modular RAG framework designed for medical question answering. It is built upon the observation that effective clinical reasoning requires structured causal inference, not merely surface-level fact retrieval. To address this, MedCoT-RAG jointly aligns document retrieval and answer generation under a unified diagnostic reasoning framework.

\begin{figure}[t]
  \centering
  \includegraphics[width=0.9\linewidth]{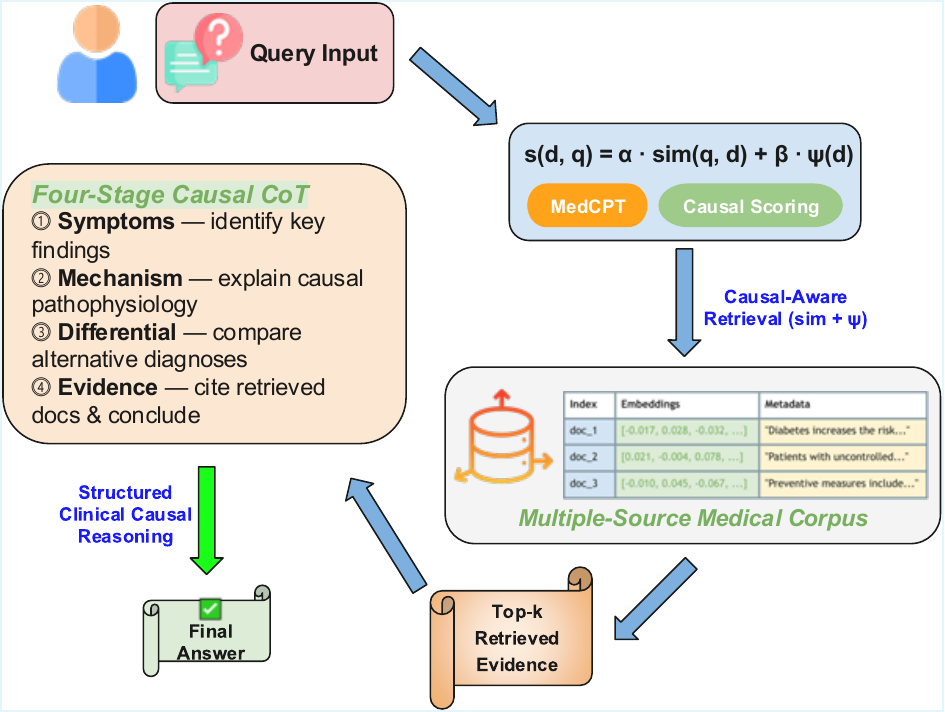}
  \caption{MedCoT-RAG pipeline combining causal-aware retrieval and four-step causal CoT prompting for clinically coherent, causally consistent answers.}
  \label{fig:medcot-pipeline}
\end{figure}

As illustrated in Fig.~\ref{fig:medcot-pipeline}, MedCoT-RAG consists of two key components:
(1) a causal-aware retrieval module that selects clinically salient documents based on both semantic relevance and causal reasoning cues, and 
(2) a structured prompting scheme that guides the LLM through step-by-step causal reasoning aligned with clinical workflows.

\subsection{Causal-Aware Retrieval}

Many clinical questions hinge on subtle causal relationships rather than shallow lexical similarities. Traditional semantic retrieval methods often retrieve contextually similar but diagnostically irrelevant documents~\cite{wang2025healthq, khatibi2025cdf}. To overcome this, we design a causal-aware retrieval mechanism that incorporates both semantic similarity and causal relevance.

Specifically, we define the retrieval score for a document \(d\) given query \(q\) as:
\begin{equation*}
s(d, q) = \alpha \cdot \text{sim}(q, d) + \beta \cdot \psi(d),
\end{equation*}
where \( \text{sim}(q, d) \) denotes the cosine similarity between query and document embeddings, and \( \psi(d) \) represents a causal relevance score that estimates the diagnostic utility of the document. The causal score \( \psi(d) \) is computed by detecting medically relevant causal patterns in the text, such as causal operators (“leads to”, “causes”, “mediates”), treatment-action-effect relations, and mechanistic disease explanations. We implement this via a weighted keyword matching scheme, normalized by document length to avoid verbosity bias.

We initialize our document encoder with the MedCPT biomedical embedding model~\cite{jin2023medcpt} and pre-embed a multi-source medical corpus comprising PubMed abstracts, StatPearls clinical reference articles, curated medical textbooks~\cite{jin2021disease}, and Wikipedia medical pages. During retrieval, the query is optionally enhanced with clinical modifiers before being encoded. We then select the top-\(k\) documents based on the composite score \( s(d, q) \), prioritizing both relevance and interpretability for downstream reasoning.

\subsection{Structured Generation via Causal CoT}

After retrieving the top-ranked documents \( \mathcal{D}_q \), we construct a structured reasoning prompt \( P(q, \mathcal{D}_q) \) that embeds both the retrieved content and a domain-specific diagnostic schema. Unlike generic chain-of-thought (CoT) prompting~\cite{wei2022chain}, our method explicitly follows a clinically inspired reasoning workflow: 
(1) identifying clinical features,
(2) explaining underlying causal mechanisms,
(3) evaluating differential diagnoses, and 
(4) synthesizing evidence into a final decision.

This four-step causal CoT template mirrors the diagnostic reasoning process used by clinicians—from symptom recognition to mechanistic explanation and evidence-based conclusion. The prompt is expressed in natural language with structured in-context instructions, guiding the LLM toward interpretable and consistent reasoning.

Formally, the generation process follows:
\begin{equation*}
y \sim p(y \mid P(q, \mathcal{D}_q)),
\end{equation*}
where \( P(q, \mathcal{D}_q) \) integrates the query, retrieved documents, and causal reasoning schema. This setup induces an inductive bias that encourages the model to produce structured causal reasoning chains, thereby reducing hallucinations and improving clinical plausibility.

For multiple-choice questions, we include instructions to explicitly provide both a justification and a final answer choice, enabling transparent and evaluable responses. For free-form questions, the model elaborates using the four-step reasoning scaffold, improving coherence and traceability.

\subsection{Unified Causal Alignment in Retrieval and Generation}

Unlike prior medical RAG systems, MedCoT-RAG jointly optimizes both retrieval and generation under a shared causal reasoning framework. This unified design ensures that both the selected documents and the generated answers are aligned with clinical diagnostic logic. Specifically, during retrieval, MedCoT-RAG selects the top-ranked documents \( \mathcal{D}_q \subset \mathcal{D} \) by maximizing a composite score that combines semantic similarity and causal relevance:
\begin{equation*}
\sum_{d \in \mathcal{D}_q} \left[ \text{sim}(q, d) + \lambda \cdot \psi(d) \right],
\end{equation*}
where \( \text{sim}(q, d) \) measures semantic similarity between the query and document, and \( \psi(d) \) captures the document’s causal utility, reflecting its relevance to diagnostic reasoning through causal phrases or clinical mechanisms. This process favors documents that are both relevant and causally informative. Following retrieval, MedCoT-RAG generates answers using a structured causal prompt:
\begin{equation*}
p(y \mid q, \mathcal{D}_q) \propto \prod_{t=1}^{T} p(y_t \mid y_{<t}, P(q, \mathcal{D}_q)),
\end{equation*}
where \( P(q, \mathcal{D}_q) \) integrates the query and retrieved documents within the four-stage causal reasoning framework. This formulation encourages the LLM to follow a step-by-step causal reasoning path, improving both answer coherence and clinical plausibility. Together, this unified causal alignment enables MedCoT-RAG to select relevant evidence and generate structured, interpretable responses tailored to complex medical question answering.

This unified approach imposes a strong causal reasoning prior throughout the pipeline—promoting causally grounded document selection and structured, interpretable answer generation. Moreover, this design is extensible: additional components, such as causal graphs or policy-learned prompting strategies, could be integrated seamlessly into either stage.


\section{Experiments and Results}

\subsection{Experimental Setup}

We evaluate the effectiveness of MedCoT-RAG on three representative medical QA benchmarks, using the LLaMA3-8B Instruct model~\cite{grattafiori2024llama} as the backbone for all methods. Our retrieval corpus integrates four medical knowledge sources: PubMed abstracts, StatPearls clinical reference articles, medical textbooks curated by Jin et al.~\cite{jin2021disease}, and Wikipedia. This multi-source corpus mirrors prior systems such as MedRAG~\cite{zhao2025medrag} and RAG$^2$~\cite{sohn2024rationale}, while extending them with newer resources like StatPearls, which support clinical decision-making but are underexplored in earlier RAG pipelines.

We evaluate on three datasets: MedQA-US~\cite{jin2021disease}, MMLU-Med~\cite{hendrycks2020measuring}, and BioASQ~\cite{tsatsaronis2012bioasq}. MedQA-US consists of 1,273 USMLE multiple-choice questions. MMLU-Med includes 1,089 questions from six biomedical subfields. BioASQ contains 618 binary yes/no questions from the 2019–2023 BioASQ Task B challenge; we remove gold evidence snippets to simulate open-domain settings.

All models are evaluated using strict match accuracy. We retrieve the top five documents from the pre-embedded corpus using FAISS indexing. The generator uses a maximum context length of 4096 tokens and generates up to 256 tokens. The same model and decoding settings are applied across all variants for fair comparison.

\subsection{Results and Analysis}

Table~\ref{tab:main_results} summarizes the accuracy of all methods across the three benchmarks. MedCoT-RAG achieves the highest accuracy in all cases, substantially outperforming both zero-shot prompting and retrieval-augmented baselines. On MedQA-US, our method reaches an accuracy of 70.1\%, compared to 53.3\% for the zero-shot baseline and 64.3\% for RAG$^2$, the strongest baseline. Similarly, on MMLU-Med, MedCoT-RAG achieves 74.4\%, marginally surpassing RAG$^2$ while maintaining consistent gains over other variants. On BioASQ-Y/N, which emphasizes biomedical research-style inference, our method scores 73.5\%, outperforming all other methods including RAG$^2$ and RGAR.

\begin{table}[t]
\centering
\scriptsize
\caption{Accuracy (\%) on medical QA benchmarks. MedCoT-RAG consistently outperforms both general and domain-adapted baselines.}
\label{tab:main_results}
\begin{tabular}{l
                S[table-format=2.1]
                S[table-format=2.1]
                S[table-format=2.1]}
\toprule
\textbf{Method} & \textbf{MedQA-US} & \textbf{MMLU-Med} & \textbf{BioASQ-Y/N} \\
\midrule
LLaMA3 (Zero-shot)      & 53.3 & 67.8 & 62.6 \\
Basic RAG               & 56.5 & 68.7 & 62.5 \\
MedCPT-RAG              & 54.6 & 62.5 & 62.9 \\
MedCPT-RAG + CoT        & 60.6 & 70.2 & 64.1 \\
Basic CoT               & 57.8 & 68.3 & 64.5 \\
\rowcolor{rowgreen}
\textbf{MedCoT-RAG (Ours)} & \textbf{70.1} & \textbf{74.4} & \textbf{73.5} \\
\midrule
RAG$^2$~\cite{sohn2024rationale} & 64.3 & 74.3 & 72.3 \\
RGAR~\cite{liang2025rgar}        & 58.8 & 70.2 & 70.6 \\
\bottomrule
\end{tabular}
\end{table}

These results suggest that improvements are not merely due to the retrieval of more relevant documents or the addition of chain-of-thought prompting in isolation. In fact, MedCPT-RAG alone performs slightly worse than standard RAG on MedQA, highlighting the limitation of relying purely on domain-specific embeddings. Similarly, while basic CoT prompting improves performance compared to zero-shot responses, it does not match the performance achieved by retrieval-augmented reasoning. The largest gains are observed when both components are combined under a causal reasoning framework, as in MedCoT-RAG.

To further dissect the contributions of individual components, we conduct an ablation study presented in Table~\ref{tab:ablation}. Removing either causal-aware retrieval or structured prompting leads to noticeable drops in accuracy. The full configuration—combining both causally prioritized document selection and structured diagnostic reasoning—achieves the strongest results across all tasks. For example, combining MedCPT-RAG with CoT improves MedQA accuracy to 60.6\%, but the full pipeline boosts performance to 70.1\%, indicating a synergistic effect. Beyond the quantitative improvements, qualitative inspection of model outputs suggests that MedCoT-RAG generates answers that are more clinically aligned and interpretable. For example, when answering USMLE-style diagnostic questions, MedCoT-RAG outputs typically trace a step-by-step reasoning path: identifying symptoms, hypothesizing mechanisms, ruling out alternatives, and citing relevant retrieved evidence. These structured responses closely resemble the reasoning used by human clinicians. In contrast, zero-shot and basic RAG methods often produce disconnected facts or memorized completions that lack internal coherence.

\begin{table}[t]
\centering
\scriptsize
\caption{Ablation study on MedCoT-RAG components. Values in parentheses show relative gains over Basic CoT.}
\label{tab:ablation}
\begin{tabular}{l
                S[table-format=2.1]
                S[table-format=2.1]
                S[table-format=2.1]}
\toprule
\textbf{Configuration} & \textbf{MedQA-US} & \textbf{MMLU-Med} & \textbf{BioASQ-Y/N} \\
\midrule
Basic CoT Only         & 57.8 & 68.3 & 64.5 \\
MedCPT-RAG Only        & 54.6\phantom{0} {\scriptsize (-3.2\%)} & 62.5 {\scriptsize (-8.5\%)} & 62.9 {\scriptsize (-2.5\%)} \\
MedCPT-RAG + CoT       & 60.6\phantom{0} {\scriptsize (+4.8\%)} & 70.2 {\scriptsize (+2.8\%)} & 64.1 {\scriptsize (-0.6\%)} \\
\rowcolor{rowgreen}
\textbf{MedCoT-RAG (Full)} & \textbf{70.1} {\scriptsize (+21.3\%)} & \textbf{74.4} {\scriptsize (+8.9\%)} & \textbf{73.5} {\scriptsize (+14.0\%)} \\
\bottomrule
\end{tabular}
\end{table}

Taken together, these findings highlight the importance of aligning retrieval and generation through a unified causal abstraction. Rather than relying solely on semantic similarity or heuristic reasoning formats, MedCoT-RAG is designed to reflect how clinicians structure diagnostic thought—retrieving documents with causal salience and using that knowledge to reason through evidence in a controlled and transparent manner. This not only improves answer accuracy but also provides better interpretability and safety assurance for clinical decision-support systems.

\section{Conclusion}

We introduced MedCoT-RAG, a RAG framework that integrates causal-aware document retrieval with structured chain-of-thought prompting to enhance medical question answering. By aligning both the retrieval and reasoning processes with the diagnostic logic used in clinical practice, MedCoT-RAG improves not only answer accuracy but also interpretability and clinical plausibility. Our method outperforms strong baselines across three diverse benchmarks, demonstrating the value of combining domain-specific knowledge with structured causal reasoning. These results highlight the importance of designing medical AI systems that go beyond surface-level retrieval and encourage trustworthy, transparent inference—an essential step toward safe and effective deployment in healthcare environments.

\bibliographystyle{unsrt}
\bibliography{ref_small}
\end{document}